\documentclass{article}

\PassOptionsToPackage{numbers, compress}{natbib}
\usepackage[final]{neurips_data_2022}

\usepackage[utf8]{inputenc} % allow utf-8 input
\usepackage[T1]{fontenc}    % use 8-bit T1 fonts
\usepackage{hyperref}
\hypersetup{
    colorlinks,
    linkcolor={red!50!black},
    citecolor={green!90!yellow},
    urlcolor={blue!80!black}
}

\usepackage{url}            % simple URL typesetting
\usepackage{booktabs}       % professional-quality tables
\usepackage{amsfonts}       % blackboard math symbols
\usepackage{nicefrac}       % compact symbols for 1/2, etc.
\usepackage{microtype}      % microtypography
\usepackage{xcolor}         % colors

\usepackage{sidecap, caption}

\usepackage{hyperref}
\usepackage{url}

\usepackage{babel}
\usepackage[font=small,labelfont=bf]{caption}

\usepackage{wrapfig}

\usepackage{times}
\usepackage{graphicx}
\usepackage{amsmath}
\usepackage{amssymb}
\usepackage{array}
\newcolumntype{?}{!{\vrule width 1.3pt}}
\usepackage{multirow}
\usepackage{makecell}
\usepackage{diagbox}

\usepackage{algorithm}
\usepackage{algorithmic}

\def\ie{i.e.,~}

\def\eg{e.g.,~}
\def\etal{et al.~}
\title{Complementary datasets to COCO \\ for object detection}

\author{%
  Ali Borji \\ % \thanks{.}
  Quintic AI, \\
  San Francisco, CA,\\
  \texttt{aliborji@gmail.com} \\
}

\begin{document}

\maketitle

% lots of stuff borrowd from other overleaf project 
% https://www.overleaf.com/project/603ecce26f90197c21c761e9

%%%%%%%%% ABSTRACT
\begin{abstract}
\vspace{-5pt}

% Add results / cite the DETR paper from FB!

For nearly a decade, the COCO dataset has been the central test bed of research in object detection. According to the recent benchmarks, however, it seems that performance on this dataset has started to saturate. One possible reason can be that perhaps it is not large enough for training deep models. To address this limitation, here we introduce two complementary datasets to COCO: i) COCO\_OI, composed of images from COCO and OpenImages (from their 80 classes in common) with 1,418,978 training bounding boxes over 380,111 images, and 41,893 validation bounding boxes over 18,299 images, and ii) ObjectNet\_D containing objects in daily life situations (originally created for object recognition known as ObjectNet; 29 categories in common with COCO). The latter can be used to test the generalization ability of object detectors. We evaluate some models on these datasets and pinpoint the source of errors. We encourage the community to utilize these datasets for training and testing object detection models. Code and data is available at \url{https://github.com/aliborji/COCO_OI}.

\end{abstract}

% \vspace{-5pt}
\section{Introduction} % and Motivation}

Object recognition is believed, although debatable, to be solved in computer vision witnessed by the ``superhuman'' performance of the state-of-the-art (SOTA) models ($\sim$3\% \emph{vs} $\sim$5\% top-5 error rate human \emph{vs} machine on ImageNet, see~\cite{he2016deep,dosovitskiy2020image,russakovsky2015ImageNet}).
Unlike object recognition, however, object detection
remains largely unsolved and models perform far below the theoretical upper bound~\cite{borji2019empirical}. The best performance on \emph{COCOval2017} and \emph{COCOtest-dev2017} are 63.2\% and 63.3\%, respectively\footnote{\url{https://paperswithcode.com/task/object-detection.}}.

A critical concern is that maybe detection datasets are not big enough to capture variations in object size\footnote{The median scale of the object relative to the image in ImageNet \emph{vs.} COCO is 554 and 106, respectively. Therefore, most object instances in COCO are smaller than 1\% of the image area~\citep{singh2018analysis}.}, viewpoint, occlusion, and spatial relationships among objects. In other words, scaling object detection seems to be much more challenging compared to scaling object recognition.

The bulk of research on object detection has been primarily focused on developing network architectures or designing new loss functions; less attention has been devoted to data. A notable exception is OpenImages dataset. Despite being orders of magnitude larger than COCO~\citep{lin2014microsoft}, however, OpenImages~\cite{kuznetsova2018open} has been less adopted. Motivated by studies (\eg~\cite{dosovitskiy2020image,kolesnikov2019big,radford2021learning}) that have used very large datasets (\eg JFT-300M~\cite{sun2017revisiting}) to achieve SOTA accuracy in object recognition (about 90\% top-1 acc; 10\% improvement over models trained on ImageNet), here we extend the COCO dataset building on top of its ecosystem. Our approach is orthogonal to works that have employed data augmentation to increase data (\eg~\cite{zhang2019bag,ghiasi2020simple}). We construct a new validation set to evaluate detectors and diagnose model errors. It is important to have multiple evaluation sets with different characteristics to gauge generalization and robustness of object detectors.

\section{Complementary datasets to COCO}

\begin{figure}[t]
\centering
  \includegraphics[width=.7\linewidth]{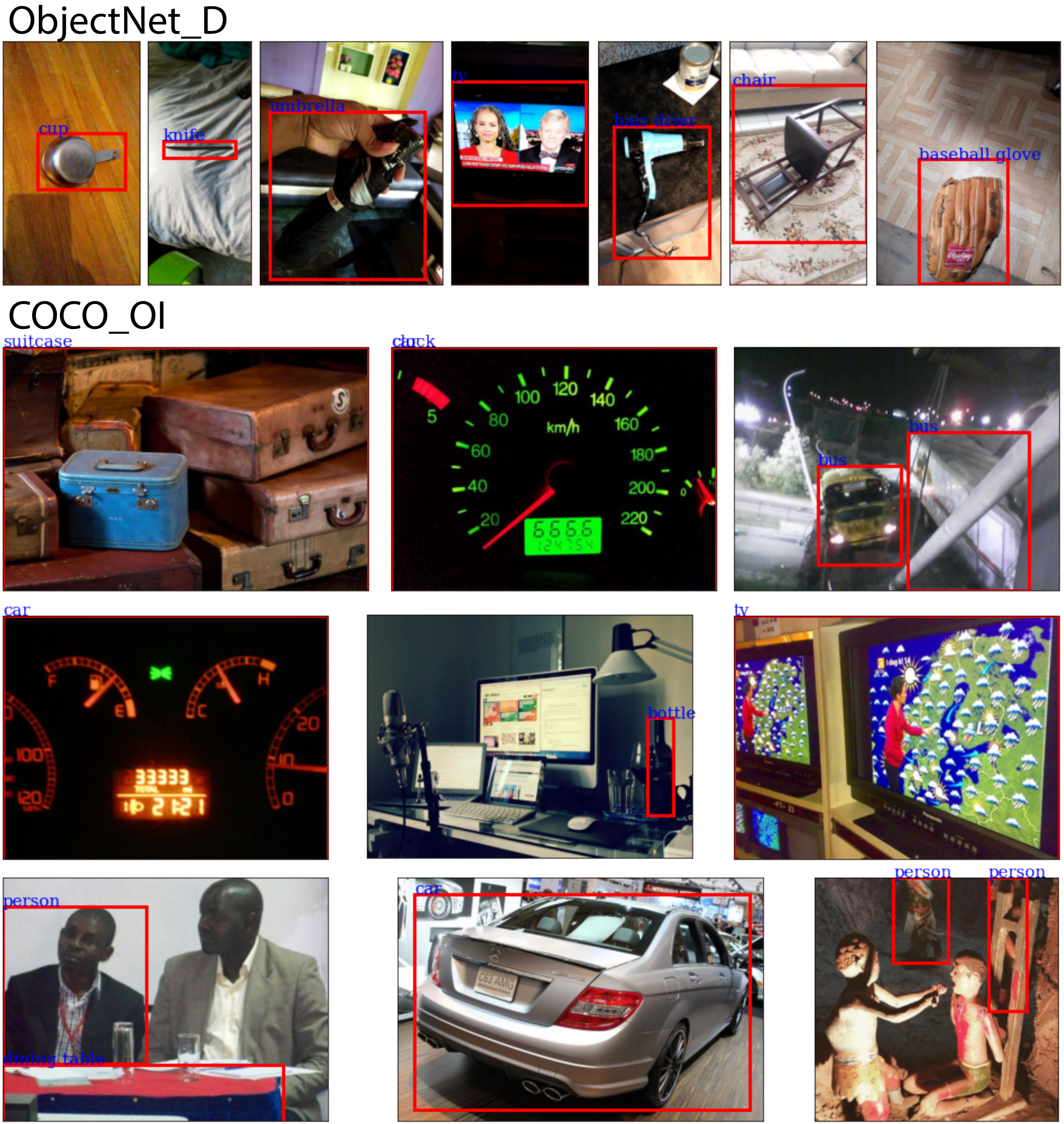}
    \caption{{Samples from validation sets of ObjectNet\_D and COCO\_OI.}} %The 
  \label{fig:new_samples}
\end{figure}

% [All 80 classes are in common between the two datasets albeit with different names.] 
\subsection{COCO\_OI}
We selected images from all 80 categories of OpenImages that are in common with COCO, except \texttt{person}, \texttt{car}, \texttt{chair} classes which are 3 most frequent classes in COCO (Fig.~\ref{fig:newHists}; Appx~\ref{stats}). This is to a) avoid making the dataset highly imbalanced\footnote{When discarding some object classes from OpenImages, any image with any object instance from these classes, even though it had some objects from other classes of interest, was discarded.}, and b) keep the number of samples and computation for model training under control. Some classes from OpenImages fall under the same COCO category (\eg \texttt{bear}, \texttt{brown bear}, \texttt{polar bear} all are mapped to the \texttt{bear} class in COCO). The class mappings are given in Appx~\ref{list_objects}. To keep the image size in the range of COCO image size, images from OpenImages are resized such that the larger dimension is 640 pixels while preserving the aspect ratio. The train set of COCO\_OI has 1,418,978 annotations in total across 380,111 images which is larger than \emph{COCOtrain2017} with 860K boxes and 118,287 images. From these, 261,824 images are selected from the train and test sets of OpenImages containing 558,977 boxes. As part of this dataset, we also create a validation set with 41,893 bounding boxes across 18,299 images chosen from validation set of OpenImages. See Fig.~\ref{fig:new_samples}. 
 
\subsection{ObjectNet\_D}
We annotated the object bounding boxes in the ObjectNet dataset originally introduced by Barbu~\etal~\cite{barbu2019objectnet}. This dataset contains objects in daily life situations (\url{https://objectnet.dev/}) and its images (50,000 across 313 categories) are pictured by Mechanical Turk workers using a mobile app in a variety of backgrounds, rotations, and imaging viewpoints.

Barbu~\etal reported a dramatic performance drop of the SOTA models on ObjectNet compared to their accuracy on ImageNet ($\sim$ 40-45\% drop). We revisited the Barbu~\etal's results and found that applying object recognition models to the isolated objects, rather than the entire scene as is done in the original paper, leads to 20-30\% performance improvement~\cite{borji2021contemplating}. 
It would be interesting to see how well object detectors perform on this dataset. In total, 5875 bounding boxes across 5875 images (one annotated object per image) from 29 categories in common with COCO are annotated. Please see Appx~\ref{stats} for more details on this dataset.

% \vspace{-8pt}
\section{Model performance and error analysis}
\vspace{-5pt}
We tested two SOTA models, Efficientdet~\cite{tan2019efficientdet}\footnote{efficientdet-d7.pth from \scriptsize{\url{https://github.com/zylo117/Yet-Another-EfficientDet-Pytorch}}} and DetectoRS~\cite{cai2018cascade} on the new validation sets. Models have been trained on COCO dataset.
% (see paperswithcode.com). 
% Performance of these models (trained on COCO) on the validation sets of the three datasets are shown in . 
As you can observe in Table~\ref{tab:new_model_perf}, both models perform lower on ObjectNet\_D and COCO\_OI datasets compared to COCO. In particular, over ObjectNet\_D, performance has been severely degraded partially because objects in this dataset are pictured in various, and sometimes odd, viewpoints and backgrounds (Fig.~\ref{fig:new_samples}). Also, AP$_S$ and AP$_M$ on this dataset are both very low since almost all objects in ObjectNet\_D are large, per the object size definitions in COCO.

\begin{table}
\begin{center}

\renewcommand{\tabcolsep}{7pt}
\renewcommand{\arraystretch}{1.1}
\begin{tabular}{l|l|ccc|ccc}
% \hline
% \textbf {Model} & \multicolumn{3}{c|} {\textbf{AP-Box}} & 
% \multicolumn{3}{c} {\textbf{AP-Mask}} \\
% \hline 
Dataset & Model & AP & AP$_{50}$ & AP$_{75}$ & AP$_S$ & AP$_M$ & AP$_L$  \\

\hline\hline
{\bf COCO} & Efficientdet & 
 51.2 & 70.4 & 55.1 & 35.7 & 55.8 & 64.8 \\
& DetectoRS & 49.1 & 67.7 & 53.4 & 29.9 & 53 & 65.2  \\

\hline
{\bf COCO\_OI} & Efficientdet & 
44 & 52.7 & 45.7 & 9 & 23.6 & 51.5 \\
& DetectoRS &  51.5 & 51 & 45.1 & 11.4 & 25.5 & 50.9 \\

\hline
{\bf ObjectNet\_D} & Efficientdet &
  23.9 & 41.6 & 24.5 & 0 & 5.7 & 24.7 \\
& DetectoRS & 20.3 & 37.8 & 19.6 & 0 & 4.7 & 21.8 \\

\end{tabular}

\end{center}
% \vspace{5pt}
\caption{Model performance over the validation sets of datasets.}
\label{tab:new_model_perf}
\end{table}

\begin{figure}[t]
\centering
  \includegraphics[width=.9\linewidth]{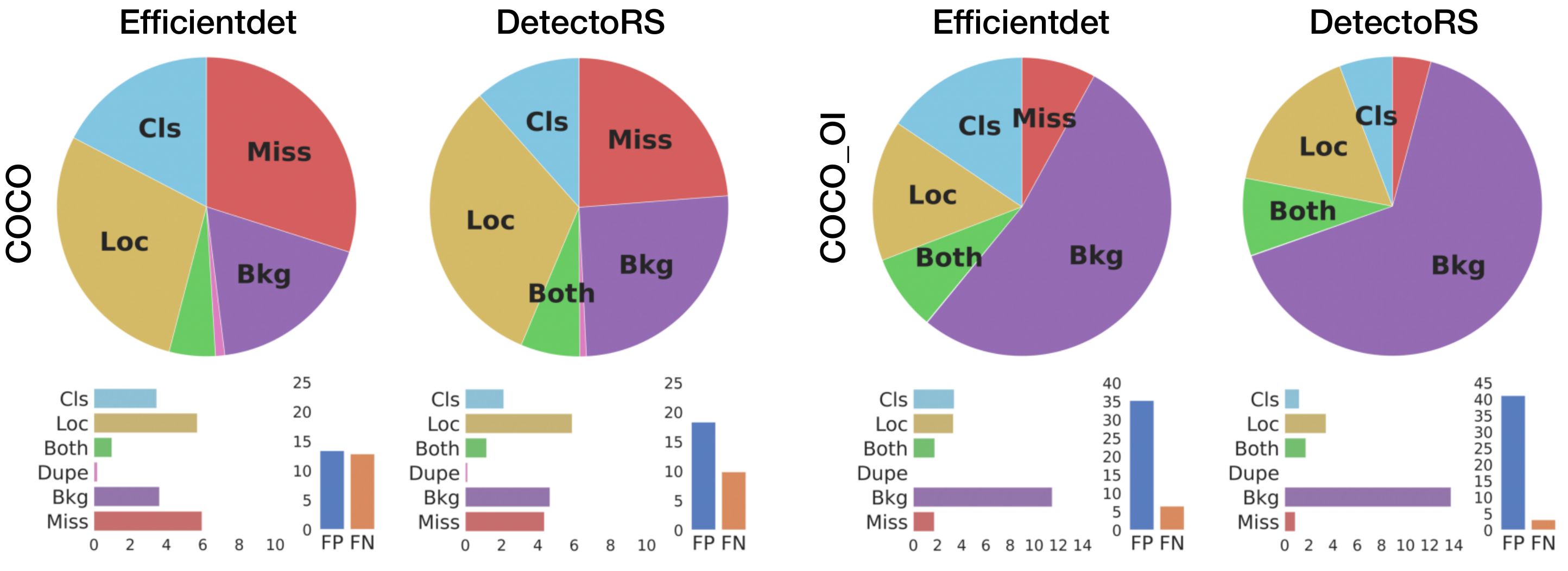}
    \caption{Error analysis over validation sets of COCO and COCO\_OI datasets.} 
  \label{fig:new_errors}
  
\end{figure}

To examine whether models behave similarly, we compare their error patterns over COCO and COCO\_OI validation sets. To this end, we employ the recently proposed error diagnosis and analysis tool known as TIDE by Bolya \etal~\cite{bolyatide}. According to Fig.~\ref{fig:new_errors}, both models behave the same on each dataset. Their behavior across datasets, however, is drastically different. On COCO, errors are almost equally distributed over 4 major error types (\texttt{Loc}, \texttt{Cls}, \texttt{Bkg}, and \texttt{Miss}), whereas over COCO\_OI, \texttt{Bkg} (\ie classifying background as an object) significantly outweighs other error types. This result signifies the impact of distribution shift on models. %[elaborate]
% [[more junk here result signifies the impact of distribution shift on models.]]

% \vspace{-10pt}
\section{Conclusion}
Object detection models have come a long way but performance is still low compared to other vision tasks such as object recognition. Heavy emphasis on the COCO dataset for model building has increased the risk of overfitting to this dataset, as is the case over ImageNet~\cite{recht2019ImageNet}. Lower performance of object detectors on ObjectNet\_D dataset (Table~\ref{tab:new_model_perf}) indicates that, similar to recognition models, object detectors also suffer from domain shift. 

We foresee the following directions for future research in this area: b) employing COCO\_OI for training models and comparing the results with models trained on COCO, and c) testing models over validation sets of the datasets, in particular ObjectNet\_D dataset, and c) using the proposed techniques to generate even larger training datasets which can open up more research opportunities for robustness, transfer learning, and semi-supervised learning in object detection,

{
\small
\bibliography{egbib}
\bibliographystyle{plain}
}

\clearpage

\section{List of object categories in new datasets}
\label{list_objects}

The list of object categories present in the COCO along with their corresponding classes in the OpenImages dataset is given below. The mapping between COCO to ObjectNet$\_D$ datasets is also presented. Note that some classes in COCO correspond to more than one class in the destination dataset (\ie one to many relationship). 
% Also, notice that images without any of these object categories were automatically discarded.

\

{\bf COCO to OpenImages category mapping:}
\begin{verbatim}
COCO_to_OI_dict = {'Baseball glove' : 'Baseball glove',
'Parking meter' : 'Parking meter', 
'Skateboard' : 'Skateboard',
'Car' : 'Car', 
'Cup' : ['coffee cup', 'Measuring cup'],
'Toilet' : 'Toilet',
'bicycle' : 'bicycle',
'sink' : 'sink',
'cow': ['Cattle', 'Bull'],
'cat': 'cat',
'Motorcycle' : 'Motorcycle',
'Bed' : 'Bed', 
'boat' : 'boat',
'Cake' : 'Cake',
'refrigerator' : 'refrigerator',
'Airplane' : 'Airplane',
'Carrot' : 'Carrot',
'Stop sign' : 'Stop sign',
'Zebra' : 'Zebra',
'Tie' : 'Tie',
'Pizza' : 'Pizza',
'Banana' : 'Banana',
'Traffic light' : 'Traffic light',
'Tennis racket' : 'Tennis racket',
'Microwave' : 'Microwave oven',
'Apple': 'Apple',
'donut': 'Doughnut', 
'Bear': ['Bear', 'Brown bear', 'Polar bear'], 
'Teddy bear' : 'Teddy bear', 
'clock': ['Alarm clock', 'Clock', 'Digital clock'], 
'sports ball' : ['Ball', 'Football', 'Volleyball' , 'Tennis Ball',  
                      'Rugby ball', 'golf ball',  'Cricket ball'], 
'horse': 'horse',
'Hair drier': 'Hair dryer', 
'Fire hydrant': 'Fire hydrant',
'Bus': 'Bus',
'Dining Table' : ['Table', 'Kitchen & dining room table'], 
'Couch': ['Couch', 'Studio couch'], 
'mouse': 'computer mouse',
'Remote': 'Remote control',
'Bench': 'Bench',
'Scissors': 'Scissors',
'Truck': 'Truck',
'Wine glass': 'Wine glass',
'Oven': 'Oven',
'Backpack': 'Backpack',
'Bird': 'Bird',
'Handbag': 'Handbag',
'Sheep': 'Sheep',
'Spoon': 'Spoon',
'Tv':'Television',
'Frisbee' : 'Flying disc',
'Keyboard' : 'computer Keyboard',
'Sandwich' : ['Sandwich', 'Submarine sandwich'],
'Cell phone' : 'Mobile phone',
'Giraffe' : 'Giraffe',
'Chair' : 'Chair',
'Bottle' : 'Bottle',
'Potted plant' : 'Houseplant', 
'Broccoli' : 'Broccoli',
'Umbrella' : 'Umbrella',
'Orange' : 'Orange',
'Hot dog' : 'Hot dog',
'Knife' : 'kitchen knife', 
'Vase' : 'Vase',
'Surfboard' : 'Surfboard',
'Toaster' : 'Toaster',
'Bowl' : ['bowl', 'Mixing bowl'],
'Suitcase' : 'Suitcase',
'Fork' : 'Fork',
'Skis' : 'Ski',
'Book' : 'Book',
'Kite' : 'Kite',
'Person' : 'Person',
'Dog' : 'Dog',
'Laptop' : 'Laptop',
'Elephant' : 'Elephant',
'Toothbrush' : 'Toothbrush',
'Baseball bat' : 'Baseball bat',
'Snowboard' : 'Snowboard',
'Train' : 'Train'}
\end{verbatim}

\vspace{20pt}

{\bf COCO to ObjectNet\_D category mapping:}
\begin{verbatim}
COCO_to_ObjNet_dict = {'Baseball glove' : 'baseball_glove',
'Cup' : ['drinking_cup', 'measuring_cup'],
'bicycle' : 'bicycle',
'Tie' : 'tie',
'Banana' : 'banana',
'Tennis racket' : 'tennis_racket',
'Microwave' : 'microwave',
'clock': 'alarm_clock',
'Hair drier': 'hair_dryer', 
'Dining Table' : 'coffee_table',
'mouse': 'computer_mouse',
'Remote': 'remote_control',
'Bench': 'bench',
'Backpack': 'backpack',
'Tv':'Tv',
'Keyboard' : 'keyboard',
'Cell phone' : 'cellphone',
'Chair' : 'chair',
'Bottle' : ['water_bottle', 'cooking_oil_bottle', 'beer_bottle', 'wine_bottle'],
'Umbrella' : 'Umbrella',
'Orange' : 'Orange',
'Knife' : ['butchers_knife' , 'bread_knife'],
'Vase' : 'Vase',
'Toaster' : 'Toaster',
'Bowl' : ['mixing_salad_bowl', 'soup_bowl'],
'Suitcase' : 'briefcase',
'Book' : 'book_closed',
'Laptop' : 'laptop_open',
'Baseball bat' : 'Baseball_bat'}
\end{verbatim}

\section{Dataset statistics}
\label{stats}

\begin{figure*}[htbp]
\begin{center}
  \includegraphics[width=1.6\linewidth,angle=90]{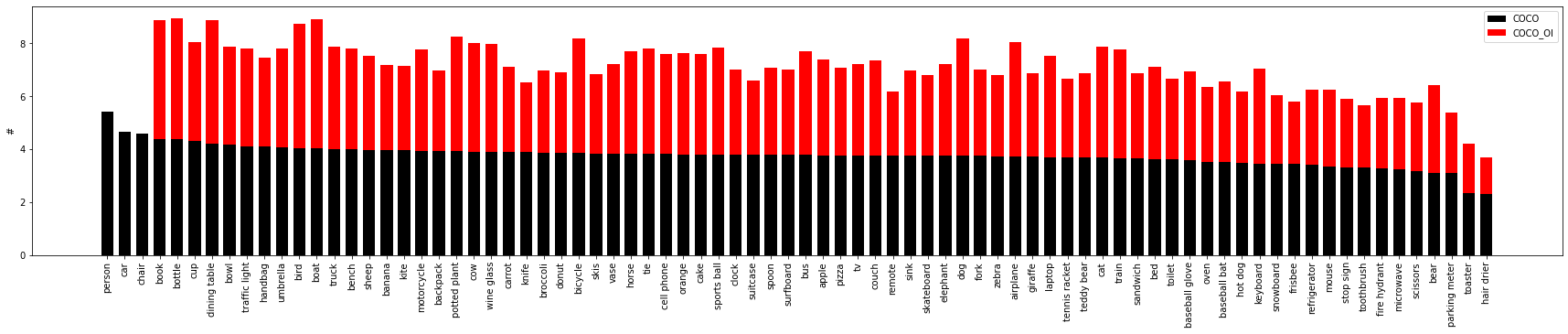}
\end{center}
 \vspace{3pt}
  \caption{Distribution of the number of objects per category in training sets of COCO and COCO\_OI datasets.}
\label{fig:newHists}
\vspace{-5pt}
\end{figure*}

Some statistics of the proposed datasets are shown here. In building $COCO\_OI$ dataset, Fig.~\ref{fig:newHists}, we have avoided to include samples from Person, Car, and Chair categories from the OpenImages dataset. Including these categories will make the $COCO\_OI$ dataset highly imbalanced. Our code for creating this dataset (\href{https://github.com/aliborji/COCO_OI}){link}, however, is very general and allows discarding arbitrary object categories to make a more balanced dataset.

\begin{figure*}[htbp]
\begin{center}
  \includegraphics[width=1.6\linewidth,angle=90]{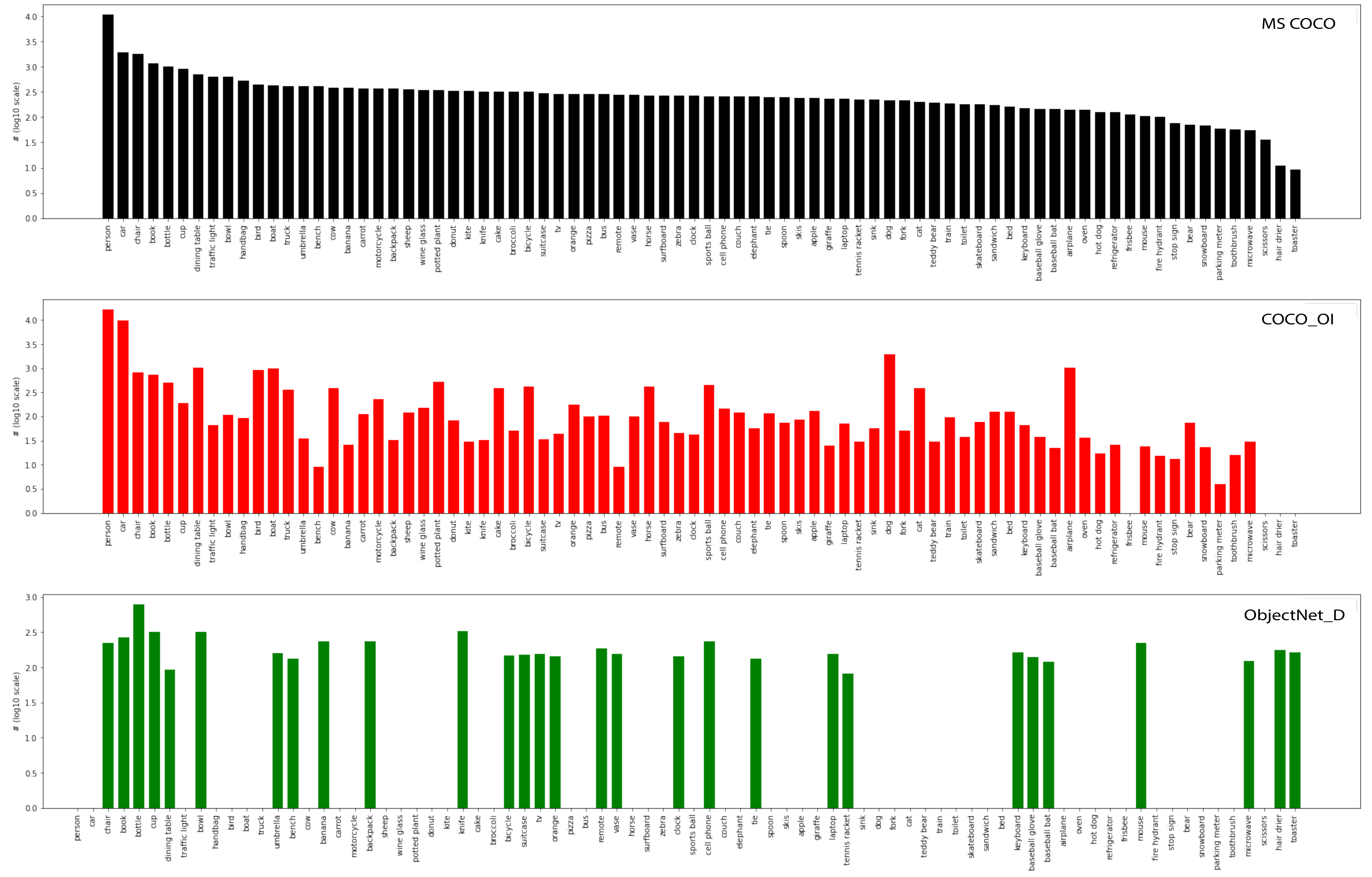}
\end{center}
  \caption{Distribution of the number of objects across the validation sets of the used datasets.}
\label{fig:newHists2}
\vspace{-5pt}
\end{figure*}

\begin{figure*}[htbp]
\begin{center}
  \includegraphics[width=1.5\linewidth,angle=90]{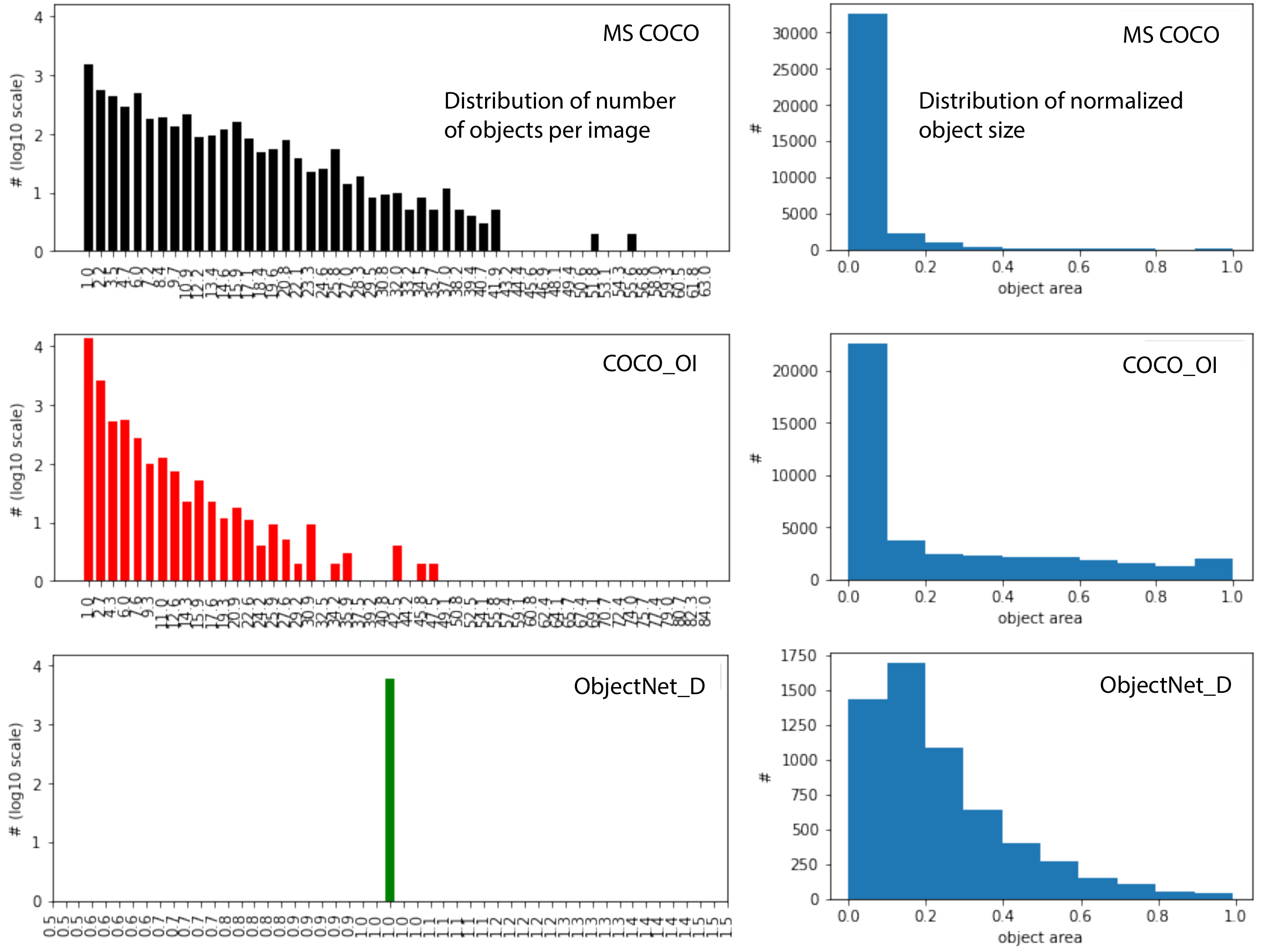}
\end{center}
  \caption{Distribution of the number of objects and normalized object size across the used datasets.}
\label{fig:newStats}
\vspace{-5pt}
\end{figure*}

\subsection{Details on ObjectNet\_D dataset}

ObjectNet dataset~\cite{barbu2019objectnet}, is build with the purpose of having less bias than other recognition datasets. It consists of indoor objects that are available to many people, are mobile, are not too large, too small, fragile or dangerous. This dataset is supposed to be used solely as a test set and comes with a licence that disallows the researchers to finetune models on it.
% \footnote{Unlike some benchmarks (\eg ImageNet) that hide their test images, images in ObjectNet dataset will be publicly available. See~\url{http://objectnet.dev}.}. 

ObjectNet images are pictured by Mechanical Turk workers using a mobile app in a variety of backgrounds, rotations, and imaging viewpoints. ObjectNet contains 50,000 images across 313 categories, out of which 113 are in common with ImageNet categories. Astonishingly, Barbu~\etal found that the state of the art object recognition models
% \footnote{They mean object recognizers!} 
perform drastically lower on ObjectNet compared to their performance on ImageNet (about 40-45\% drop).

\begin{figure*}[htbp]
\begin{center}
  \includegraphics[width=\linewidth]{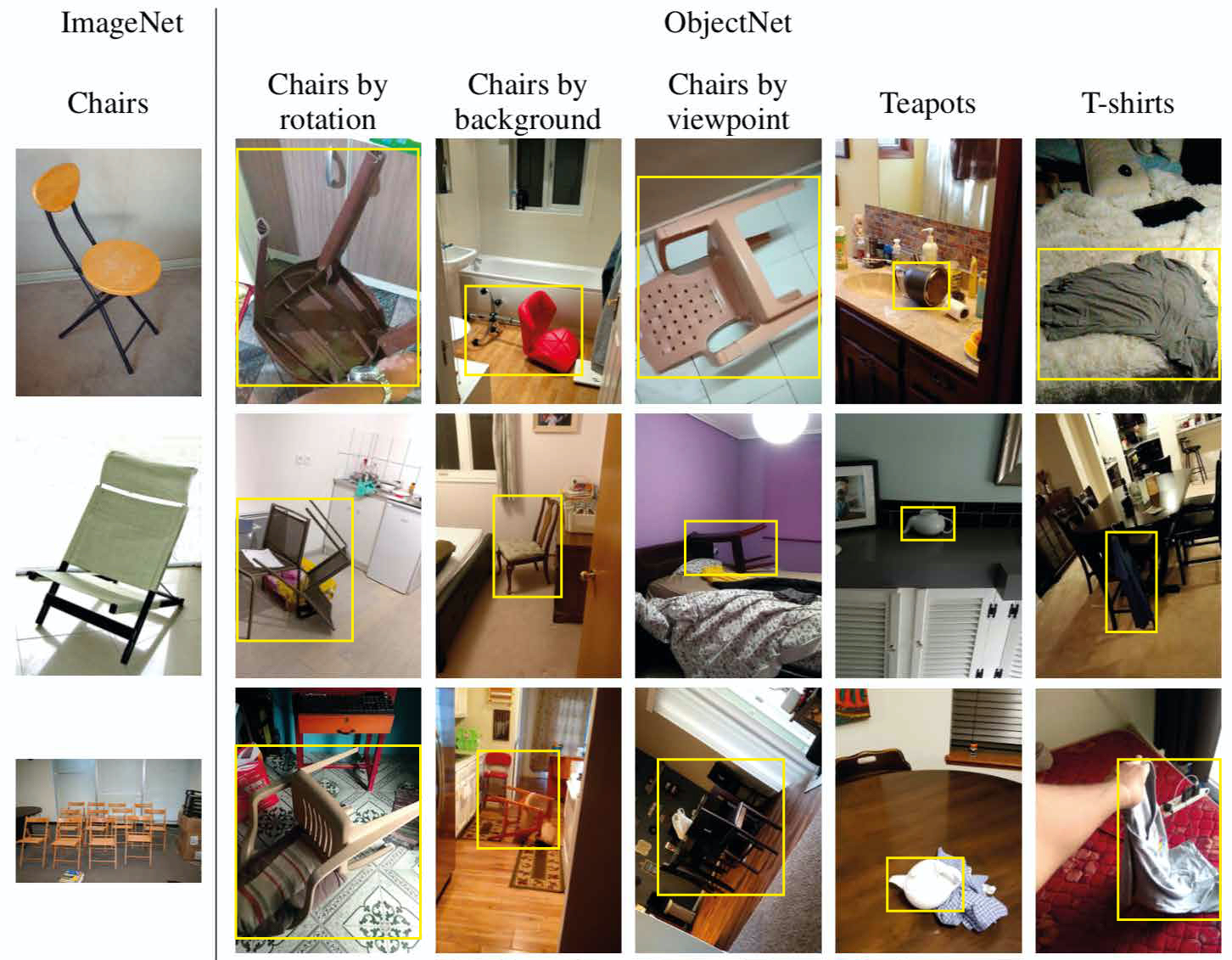}
\end{center}
  \caption{Sample images from the ObjectNet dataset along with our manually annotated object bounding boxes from \texttt{Chairs}, \texttt{Teapots} and \texttt{T-shirts} categories. The leftmost column shows three chair examples from the ImageNet dataset. ImageNet scenes often have a single isolated object in them whereas images in the ObjectNet dataset contain multiple objects. Further, ObjectNet objects cover a wider range of variation in contrast, rotation, scale, and occlusion compared to ImageNet objects (See arguments in~\cite{barbu2019objectnet}). In total, we annotated 18,574 images across 113 categories in common between the two datasets. This figure is modified from Figure 2 in~\cite{barbu2019objectnet}.}
\label{fig:box}
\end{figure*}

The 113 object categories in the ObjectNet dataset, overlapped with the ImageNet, contain 18,574 images in total. On this subset, the average number of images per category is 164.4 ({\em min}=55, {\em max}=284). We drew
a bounding box around the object corresponding to the category label of each image. If there were multiple nearby objects from the same category (\eg chairs around a table), we tried to include all of them in the bounding box. Some example scenes and their corresponding bounding boxes are given in Fig.~\ref{fig:box}.

\begin{figure*}[htbp]
\begin{center}
  \includegraphics[width=.6\linewidth]{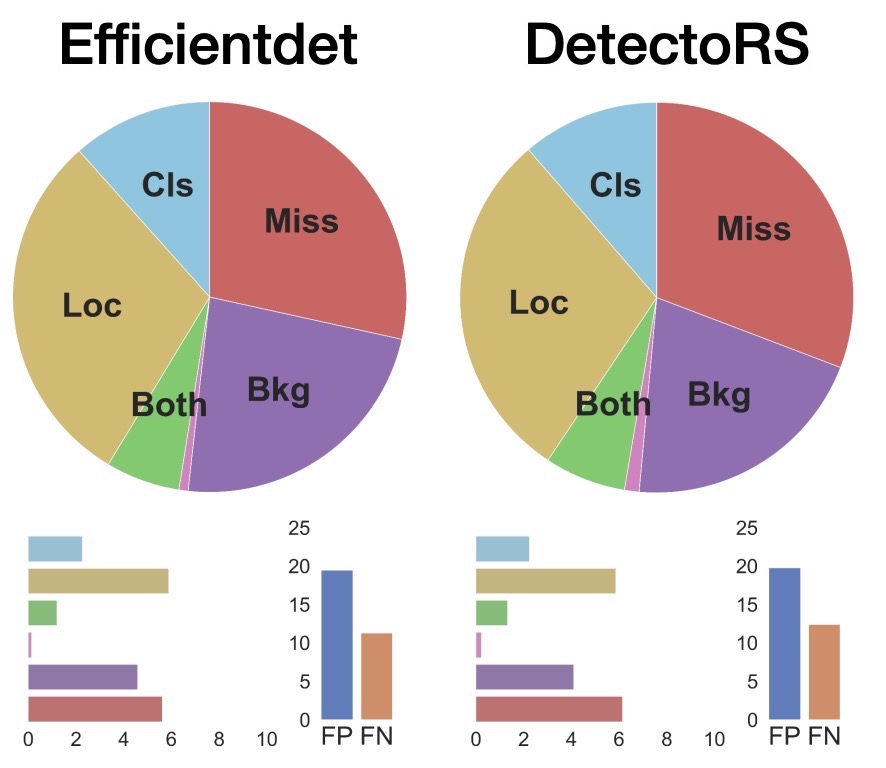}
\end{center}
  \caption{Error analysis over the ObjectNet\_D dataset.}
\label{fig:annotation_errorsXX}
\end{figure*}

\clearpage

% \clearpage
\section{Dataset licenses}
\label{licenses}

Below we list each dataset’s license, as provided either in the paper proposing the dataset or on the dataset website. 

\begin{enumerate}
    \item MSCOCO:  Creative Commons Attribution 4.0 License, \url{https://cocodataset.org/#termsofuse}
    \item OpenImages: CC BY 4.0 license, the images are listed as having a CC BY 2.0 license \url{https://storage.googleapis.com/openimages/web/factsfigures.html}
    \item ObjectNet: Creative Commons Attribution 4.0 with only two additional clauses: a) ObjectNet may never be used to tune the parameters of any model, and b) Any individual images from ObjectNet may only be posted to the web including their 1 pixel red border. See \url{https://objectnet.dev/download.html}.
\end{enumerate}

\end{document}